\documentclass{article}
\usepackage{ijcai26}

\usepackage{xcolor}
\usepackage{tikz}

\usepackage{times}
\usepackage{soul}
\usepackage{url}
\usepackage[hidelinks]{hyperref}
\usepackage[utf8]{inputenc}
\usepackage[small]{caption}
\usepackage{graphicx}
\usepackage{amsmath}
\usepackage{amsthm}
\usepackage{booktabs}
\usepackage{algorithm}
\usepackage{algorithmic}
\usepackage[switch]{lineno}

\usepackage{tabularx}
\usepackage{threeparttable}
\usepackage{pifont}
\usepackage[nointegrals]{wasysym}
\usepackage[shortcuts]{extdash}
\usepackage{float}

\usetikzlibrary{shapes.geometric, arrows.meta, positioning, fit, backgrounds, calc}

\newcommand{\yes}{\CIRCLE}               
\newcommand{\partialc}{\LEFTcircle}      
\newcommand{\noo}{\Circle}               

\title{\textsc{Llars}: Enabling Domain Expert \& Developer Collaboration for LLM Prompting, Generation and Evaluation}

\author{
Philipp Steigerwald$^1$\and
Mara Stieler$^2$\and
Jennifer Burghardt$^2$\and
Eric Rudolph$^1$\and
Jens Albrecht$^1$\\
\affiliations
Technische Hochschule Nürnberg Georg Simon Ohm, Nürnberg, Germany\\
$^1$Faculty of Computer Science, Centre for Artificial Intelligence (KIZ)\\
$^2$Faculty of Social Sciences, Institute for E-Counselling
\emails
\{philipp.steigerwald, mara.stieler, jennifer.burghardt, eric.rudolph, jens.albrecht\}@th-nuernberg.de
}

\begin{document}

\maketitle

\begin{abstract}
We demonstrate \textsc{Llars} (LLM Assisted Research System), an open-source platform that bridges the gap between domain experts and developers for building LLM-based systems.
It integrates three tightly connected modules into an end-to-end pipeline:
\textbf{Collaborative Prompt Engineering} for real-time co-authoring with version control and instant LLM testing,
\textbf{Batch Generation} for configurable output production across user-selected prompts\,$\times$\,models\,$\times$\,data with cost control and
\textbf{Hybrid Evaluation} where human and LLM evaluators jointly assess outputs through diverse assessment methods, with live agreement metrics and provenance analysis to identify the best model--prompt combination for a given use case.
New prompts and models are automatically available for batch generation and completed batches can be turned into evaluation scenarios with a single click.
Interviews with six domain experts and three developers in online counselling confirmed that \textsc{Llars} feels intuitive, saves considerable time by keeping everything in one place and makes interdisciplinary collaboration seamless.

\vspace{0.3em}
\noindent\textbf{Source code:} \url{github.com/th-nuernberg/llars} 
\end{abstract}

\section{Introduction}

\definecolor{clrPrompt}{HTML}{B0CA97}
\definecolor{clrBatch}{HTML}{88C4C8}
\definecolor{clrEval}{HTML}{D1BC8A}
\definecolor{clrOutcome}{HTML}{E8A087}

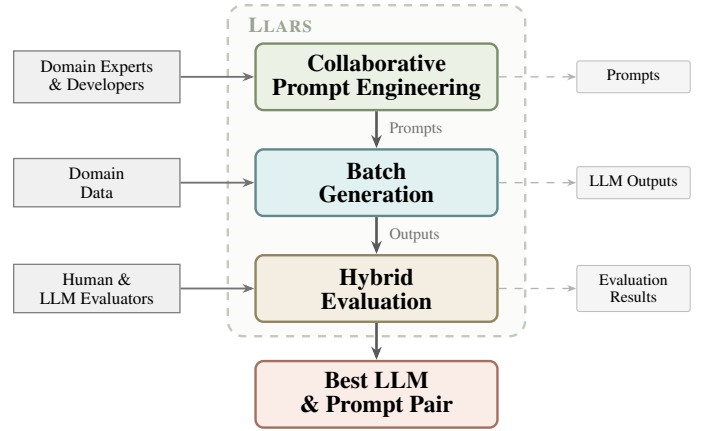
\begin{figure}[t]
\centering
\begin{tikzpicture}[
    module/.style={
        rectangle, minimum width=3.2cm, minimum height=0.88cm,
        align=center, font=\footnotesize\bfseries,
        rounded corners=3pt, line width=0.9pt
    },
    actor/.style={
        rectangle, draw=black!50, fill=black!6,
        minimum width=2.2cm, minimum height=0.62cm,
        align=center, font=\fontsize{6.5}{8}\selectfont,
        line width=0.5pt
    },
    export/.style={
        rectangle, draw=black!25, fill=black!4,
        minimum width=1.5cm, minimum height=0.42cm,
        align=center, font=\fontsize{6}{7.5}\selectfont,
        rounded corners=1pt
    },
    outcome/.style={
        rectangle, draw=clrOutcome!70!black, fill=clrOutcome!20,
        minimum width=3.2cm, minimum height=0.88cm,
        align=center, font=\footnotesize\bfseries,
        rounded corners=3pt, line width=0.9pt
    },
    arrow/.style={-{Stealth[length=4.5pt]}, line width=0.7pt, black!55},
    pipearrow/.style={-{Stealth[length=5pt, width=4pt]}, line width=1.0pt, black!65},
    dasharrow/.style={-{Stealth[length=3.5pt]}, line width=0.5pt, black!30, dashed},
    lbl/.style={font=\fontsize{6}{7.5}\selectfont, text=black!55},
]

\node[module, draw=clrPrompt!70!black, fill=clrPrompt!25] (pe) at (0, 0)
    {Collaborative\\[-1pt]Prompt Engineering};

\node[module, draw=clrBatch!70!black, fill=clrBatch!25] (bg) at (0, -1.4)
    {Batch\\[-1pt]Generation};

\node[module, draw=clrEval!70!black, fill=clrEval!25] (ev) at (0, -2.8)
    {Hybrid\\[-1pt]Evaluation};

\node[actor] (dev) at (-3.7, 0) {Domain Experts\\[-1pt]\& Developers};
\node[actor] (data) at (-3.7, -1.4) {Domain\\[-1pt]Data};
\node[actor] (llmeval) at (-3.7, -2.8) {Human \&\\[-1pt]LLM Evaluators};

\draw[arrow] (dev.east) -- (pe.west);
\draw[arrow] (data.east) -- (bg.west);
\draw[arrow] (llmeval.east) -- (ev.west);

\draw[pipearrow] (pe.south) -- node[lbl, right, xshift=1pt] {Prompts} (bg.north);
\draw[pipearrow] (bg.south) -- node[lbl, right, xshift=1pt] {Outputs} (ev.north);

\node[export] (exp1) at (3.4, 0) {Prompts};
\node[export] (exp2) at (3.4, -1.4) {LLM Outputs};
\node[export] (exp3) at (3.4, -2.8) {Evaluation\\[-1pt]Results};
\draw[dasharrow] (pe.east) -- (exp1);
\draw[dasharrow] (bg.east) -- (exp2);
\draw[dasharrow] (ev.east) -- (exp3);

\node[outcome] (outcome) at (0, -4.2) {Best LLM \\[-1pt]\& Prompt Pair};
\draw[pipearrow] (ev.south) -- (outcome.north);

\begin{scope}[on background layer]
\draw[draw=clrPrompt!60!black!40, dashed, rounded corners=6pt, line width=0.9pt,
      fill=clrPrompt!5]
      ([xshift=-10pt, yshift=14pt]pe.north -| pe.west) rectangle
      ([xshift=10pt, yshift=-5pt]ev.south -| ev.east);
\node[font=\fontsize{8}{10}\selectfont\bfseries, text=clrPrompt!50!black!60,
      anchor=north west] at ([xshift=-10pt+4pt, yshift=14pt-1pt]pe.north -| pe.west) {\textsc{Llars}};
\end{scope}

\end{tikzpicture}
\caption{\textsc{Llars} pipeline: domain experts and developers collaboratively develop prompts, generate outputs across LLMs and run hybrid evaluation with human and LLM evaluators. Each stage supports export and the pipeline yields a validated model--prompt combination.}
\label{fig:pipeline}
\end{figure}

Building LLM-based systems for sensitive domains~\cite{chen2024critical,ling2025domain}\--online counselling~\cite{stade2024behavioral}, legal analysis~\cite{katz2024bar}, medical documentation~\cite{vanveen2024clinical}\--demands both domain expertise and technical skill~\cite{zamfirescu2023johnny,schulhoff2024prompt}: practitioners must iterate prompts, select suitable models and rigorously evaluate outputs on their own data, since public benchmarks rarely transfer.
Yet domain experts and developers typically work in isolation, with prompts iterated in shared documents without version control, models tested ad hoc and outputs scattered across tools.
\textsc{Llars} (LLM Assisted Research System) is an open-source platform that unifies these three stages (Figure~\ref{fig:pipeline}).

It integrates three tightly connected modules.
\emph{Collaborative Prompt Engineering} lets domain experts and developers co-author prompts in a shared real-time editor with version control and instant LLM testing.
\emph{Batch Generation} produces the user-configured Cartesian product of prompts\,$\times$\,models\,$\times$\,data items with cost estimation and budget control.
\emph{Hybrid Evaluation} combines human and LLM evaluators in shared assessment campaigns with automated agreement statistics and provenance tracing to surface the best model--prompt pair.
New prompts, models and providers are instantly available across all modules and shareable with collaborators. Completed batches become evaluation scenarios with a single click and each module supports JSON/CSV export.

\begin{table*}[t]
\centering
\caption{Comparison of GUI-first tools for prompt engineering, batch generation and hybrid LLM output evaluation.
\yes~=~native support;\enspace \partialc~=~partial;\enspace \noo~=~not supported.\enspace
\textbf{OSS}~=~open-source software;\enspace
\textbf{Collab.}~=~simultaneous multi-user prompt editing with live synchronisation;\enspace
\textbf{Batch}~=~configurable generation over prompts\,$\times$\,models\,$\times$\,data with cost control;\enspace
\textbf{H-Eval}~=~structured human evaluation with item distribution and role management;\enspace
\textbf{LLM-E}~=~any form of automated LLM-based evaluation, including LLM-as-judge and LLM-as-evaluator;\enspace
\textbf{Stats}~=~live analytics including inter-rater reliability and provenance analysis;\enspace
\textbf{E2E}~=~integrated end-to-end pipeline.}
\label{tab:comparison}
\small
\renewcommand{\arraystretch}{1.1}
\setlength{\tabcolsep}{0pt}
\begin{tabular*}{\textwidth}{@{\extracolsep{\fill}} l c c c c c c c}
\toprule
\textbf{Tool} & \textbf{OSS} & \textbf{Collab.} & \textbf{Batch} & \textbf{H-Eval} & \textbf{LLM-E} & \textbf{Stats} & \textbf{E2E} \\
\midrule
\textbf{\textsc{Llars}}                         & \yes & \yes      & \yes      & \yes      & \yes      & \yes      & \yes \\
Agenta~\cite{agenta2024}               & \yes & \partialc & \partialc & \partialc & \yes      & \noo      & \partialc \\
ChainForge~\cite{arawjo2024chainforge} & \yes & \noo      & \yes      & \noo      & \partialc & \noo      & \partialc \\
Phoenix~\cite{phoenix2024}             & \yes & \noo      & \partialc & \partialc & \yes      & \noo      & \noo \\
Langfuse~\cite{langfuse2024}           & \yes & \noo      & \partialc & \partialc & \yes      & \partialc & \noo \\
W\&B~Weave~\cite{weave2024}            & \yes & \noo      & \partialc & \partialc & \yes      & \noo      & \noo \\
Label Studio~\cite{labelstudio2024}    & \yes & \noo      & \noo      & \yes      & \partialc & \partialc & \noo \\
Argilla~\cite{argilla2024}             & \yes & \noo      & \noo      & \yes      & \partialc & \partialc & \noo \\
LangSmith~\cite{langsmith2024}         & \noo & \noo      & \partialc & \partialc & \yes      & \noo      & \partialc \\
Braintrust~\cite{braintrust2024}       & \noo & \noo      & \partialc & \partialc & \yes      & \noo      & \partialc \\
Maxim~\cite{maxim2024}                 & \noo & \noo      & \partialc & \partialc & \yes      & \noo      & \partialc \\
Vellum~\cite{vellum2024}               & \noo & \noo      & \partialc & \partialc & \yes      & \noo      & \partialc \\
\bottomrule
\end{tabular*}
\end{table*}

Table~\ref{tab:comparison} surveys twelve existing platforms along these dimensions.
Agenta~\cite{agenta2024} combines prompt versioning with LLM-as-judge scoring and A/B deployment testing but lacks real-time co-editing and structured human evaluation with inter-rater agreement.
Phoenix~\cite{phoenix2024}, Langfuse~\cite{langfuse2024} and W\&B~Weave~\cite{weave2024} provide observability dashboards with dataset experiments but focus on tracing and debugging production systems rather than systematic batch generation or coordinated multi-evaluator campaigns.
ChainForge~\cite{arawjo2024chainforge} offers a visual node-based interface for multi-model comparison with template variables but omits structured human evaluation entirely.
Label~Studio~\cite{labelstudio2024} and Argilla~\cite{argilla2024} offer dedicated human annotation workflows with agreement analytics but require externally generated outputs, disconnecting evaluation from prompt development.
LangSmith~\cite{langsmith2024}, Braintrust~\cite{braintrust2024}, Maxim~\cite{maxim2024} and Vellum~\cite{vellum2024} provide playground-style testing with dataset-level scoring but are closed-source, lack collaborative editing and omit batch generation.
\textsc{Llars} unifies collaborative prompt engineering, batch generation and hybrid human--LLM evaluation with agreement analytics in a single open-source platform.

\section{Platform Overview}

Co-designed with developers, social science researchers and practising counsellors, \textsc{Llars} runs as a containerised web application for textual data.
The three modules are tightly integrated so that prompts created in the editor flow directly into batch generation and completed batches become evaluation scenarios with a single click, keeping provenance intact.

\subsection{Collaborative Prompt Engineering}

\begin{figure}[htb]
\centering
\includegraphics[width=1.00\columnwidth]{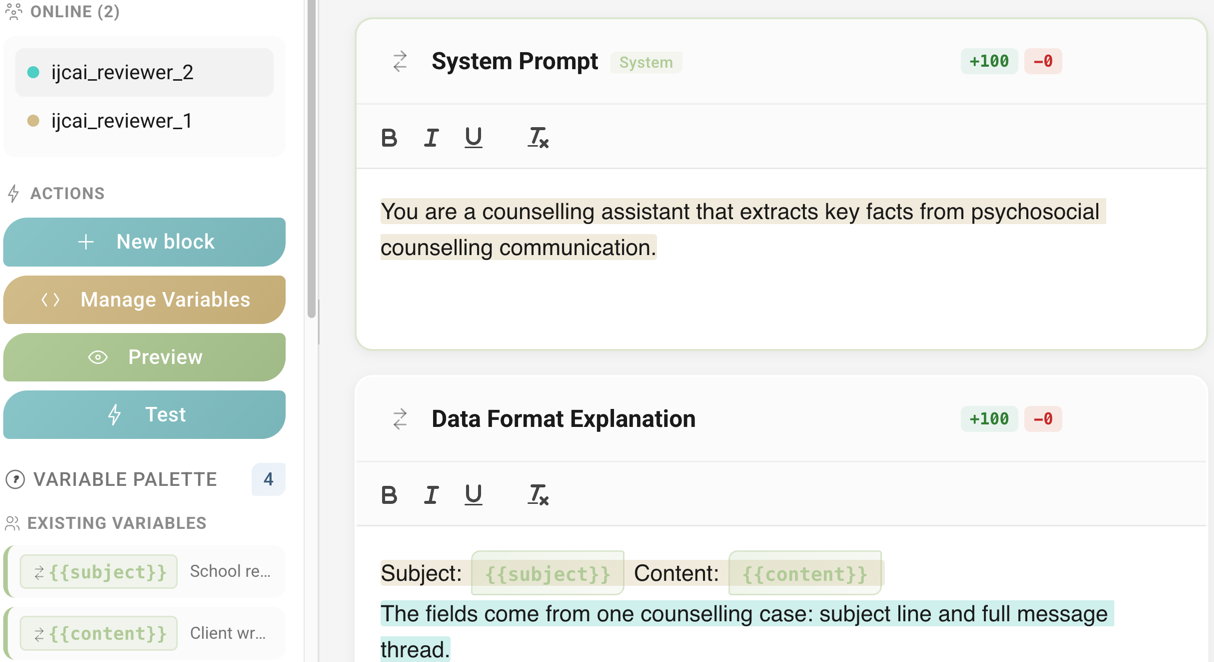}
\caption{Collaborative prompt editor with ordered blocks and template variables inserted from the Variable Palette (bottom-left).}
\label{fig:prompt-editor}
\end{figure}

The collaborative prompt editor (Figure~\ref{fig:prompt-editor}) lets every keystroke appear instantly for all connected users.
A prompt is composed of ordered blocks\--one optionally designated as system prompt, the rest concatenated into the user prompt\--each maintaining its own version history that tracks insertions and deletions (visible as \texttt{+100\,/\,{-}0} in Figure~\ref{fig:prompt-editor}), enabling diff comparison and rollback without affecting other blocks.

Blocks may contain template variables (\texttt{\{\{variable\_name\}\}}) as placeholders for external data (e.g.\ an entire email thread captured in a single \texttt{\{\{content\}\}} variable), keeping the surrounding instructions concise.
All variables are collected in the shared Variable Palette (Figure~\ref{fig:prompt-editor}, bottom-left) and annotated with sample values.
A single click sends the assembled prompt with sample values substituted to a selected model and streams the response back in real time.
Prompts are exportable as JSON and directly available in batch generation, where template variables are filled from the uploaded data to scale a single prompt to hundreds of outputs.
\subsection{Batch Generation}
Batch generation extends prompt testing beyond single examples by combining any set of prompts, models and data items into their Cartesian product (prompts\,$\times$\,models\,$\times$\,data items).
A generation matrix previews all combinations and estimates cost; optional budget caps pause jobs that exceed a threshold.
For example, 50~counselling email threads with two prompts and two models produce 50\,$\times$\,2\,$\times$\,2\,=\,200~outputs.
Results stream in real time, each tagged with full provenance (source item, prompt version, model, parameters, tokens and cost) and exportable as CSV or JSON.
Completed batches become evaluation scenarios in one click, preserving attribution end-to-end.

\subsection{Hybrid Evaluation}

Each evaluation is organised as a \emph{scenario}, a self-contained campaign that defines the evaluation type, assigns evaluators and distributes items.
Items are presented in randomised order and without any provenance information, so evaluators cannot tell which model or prompt produced a given output.
\textsc{Llars} supports multi-dimensional rating with configurable Likert scales, ranking into ordinal buckets or traditional ranking, categorical labelling, pairwise comparison, mail assessment and authenticity detection.
For comparing multiple outputs per input, \emph{bucket ranking} proved the most popular method in user feedback: items are sorted into ordinal categories and ranked within each bucket~\cite{miller1956magical,kiritchenko2017bws}.
For single outputs, multi-dimensional rating is more appropriate. The \emph{mail rating} preset, for instance, provides Likert scales tailored to assessing an LLM's reply within an email counselling thread.
Scenarios are created from batch outputs with a single click or via an AI-assisted Scenario Wizard.

\begin{figure}[htb]
\centering
\includegraphics[width=\linewidth]{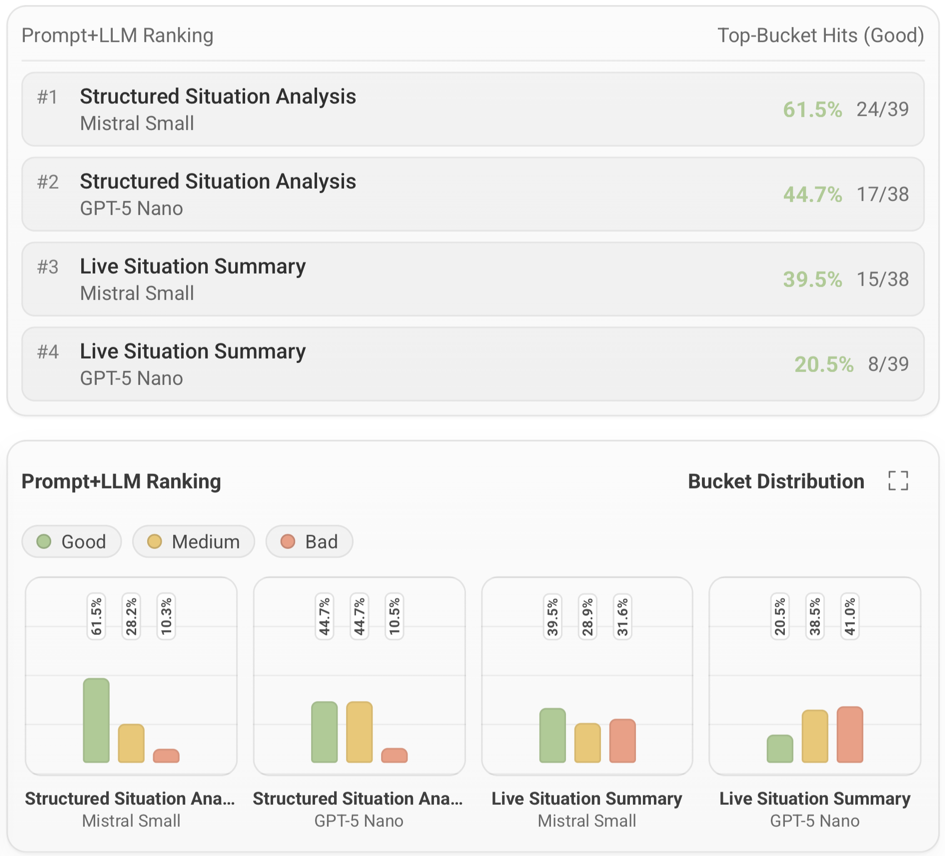}
\caption{Provenance analysis ranking model--prompt combinations by top-bucket hit rate (top) with per-combination bucket distributions (bottom).}
\label{fig:provenance}
\end{figure}

Owners control whether all items go to every evaluator or configurable subsets are distributed for faster throughput.
\textsc{Llars} treats LLM evaluators as full participants alongside humans. They receive the same items, use the same setup and their judgements enter the same agreement analysis.
Scenario owners can inspect aggregated results at any time as assessments come in, including automatically computed inter-rater reliability (e.g.\ Krippendorff's~$\alpha$), filtered by human evaluators only, LLM evaluators only or both combined.
Among the available analytics, \emph{provenance analysis} leverages each item's generating model and prompt to display the top-bucket hit rate and full bucket distribution per model--prompt pair (Figure~\ref{fig:provenance}), surfacing the best performer.

\section{Deployment and User Study}
\label{sec:counseling}

\textsc{Llars} is actively used in online counselling research: the Virtual Client project~\cite{rudolph2024virtualclient} uses it for prompt development of simulated client interactions, and CAIA~\cite{steigerwald2025caia}, an AI assistant for email counselling, relies on it for prompt development and evaluation.
A full end-to-end case study~\cite{steigerwald2025subjectline} produced 253~subject lines across 11~LLMs and had five professionals and an LLM evaluator rate them (1,518~assessments), determining how small a model could be while still meeting quality requirements.

Semi-structured interviews with six domain experts and three developers confirmed that consolidating the pipeline in one platform saves considerable time and makes interdisciplinary collaboration seamless.
Domain experts described the interface as intuitive: ``as an evaluator you receive the task and immediately know what to do.'' They also noted that testing prompts directly against an LLM ``developed a feeling for how to formulate the task to get the desired output.''
Developers valued that prompts, data and outputs remain in one place.
Both groups emphasised that working in a shared workspace ``finally enabled us to work together'' rather than passing documents back and forth between disciplines.

\section{Demo and Conclusion}

Our demo\footnote{Video: \url{https://youtu.be/3QaKouwr4gU}} walks attendees through the complete workflow, from collaborative prompt authoring through batch generation to evaluation with live agreement metrics and provenance analysis, using pre-built counselling scenarios that let attendees author prompts, trigger batch runs and evaluate outputs hands-on.

\textsc{Llars} unifies prompt engineering, batch generation and evaluation in a single open-source platform that closes the loop from prompt development to rigorous assessment.
While currently deployed in online counselling, the platform is domain-agnostic and applicable to any text-based LLM evaluation task.
Present limitations include the focus on single-turn generation rather than multi-turn conversations and the dependence of LLM-based evaluators on the underlying model's reasoning capability.
Despite these constraints, interviews confirmed that centralising all stages in one workspace saves considerable time, feels intuitive and eliminates the ``translation work'' between disciplines. Domain experts reported being able to author prompts and conduct evaluations independently without external guidance.
Beyond the current single-turn focus, future work will extend \textsc{Llars} to multi-turn conversational evaluation in which each turn is assessed in context, automated calibration of LLM evaluators against human ratings to surface systematic biases in real time and an HTTP API connecting evaluation outcomes directly to model fine-tuning pipelines, enabling closed-loop iteration from prompt development through assessment to model improvement.
Ultimately, \textsc{Llars} bridges the gap between domain expertise and technical implementation and enables interdisciplinary work.


\bibliographystyle{named}
\bibliography{llars}

@inproceedings{steigerwald2025subjectline,
  title={Comparing Large Language Models for Automated Subject Line Generation in e-Mental Health: A Performance Study},
  author={Steigerwald, Philipp and Albrecht, Jens},
  booktitle={Proceedings of the 11th International Conference on Information and Communication Technologies for Ageing Well and e-Health (ICT4AWE)},
  pages={70--77},
  year={2025},
  publisher={SciTePress},
  doi={10.5220/0013294100003938}
}

@inproceedings{steigerwald2025caia,
  title={{CAIA} in Practice: Field Evaluation of an {AI}-Assisted Support System for Text-Based Online Counselling},
  author={Steigerwald, Philipp and Bienlein, Nico and Burghardt, Jennifer and Stieler, Mara and Lehmann, Robert and Albrecht, Jens},
  booktitle={Proceedings of the IEEE International Conference on Tools with Artificial Intelligence (ICTAI)},
  year={2025},
  pages={1476--1483},
  publisher={IEEE}
}

@inproceedings{rudolph2024virtualclient,
  title={An {AI}-Based Virtual Client for Educational Role-Playing in the Training of Online Counselors},
  author={Rudolph, Eric and Engert, Natalie and Albrecht, Jens},
  booktitle={Proceedings of the 16th International Conference on Computer Supported Education (CSEDU)},
  pages={108--117},
  year={2024},
  doi={10.5220/0012690700003693}
}

@inproceedings{arawjo2024chainforge,
  title={{ChainForge}: A Visual Toolkit for Prompt Engineering and {LLM} Hypothesis Testing},
  author={Arawjo, Ian and Swoopes, Chelse and Vaithilingam, Priyan and Wattenberg, Martin and Glassman, Elena},
  booktitle={Proceedings of the 2024 CHI Conference on Human Factors in Computing Systems},
  year={2024},
  publisher={ACM},
  doi={10.1145/3613904.3642016}
}

@misc{agenta2024,
  title={Agenta: Open-Source {LLMOps} Platform for Prompt Management, Evaluation, and Observability},
  author={{Agenta AI}},
  year={2024},
  howpublished={\url{https://agenta.ai}},
  note={Accessed: 2026-02-01}
}

@misc{phoenix2024,
  title={Phoenix: Open-Source {AI} Observability and Evaluation},
  author={{Arize AI}},
  year={2024},
  howpublished={\url{https://phoenix.arize.com}},
  note={Accessed: 2026-02-01}
}

@misc{langfuse2024,
  title={Langfuse: Open-Source {LLM} Engineering Platform},
  author={{Langfuse}},
  year={2024},
  howpublished={\url{https://langfuse.com}},
  note={Accessed: 2026-02-01}
}

@misc{langsmith2024,
  title={{LangSmith}: {AI} Agent Evaluation Platform},
  author={{LangChain, Inc.}},
  year={2024},
  howpublished={\url{https://www.langchain.com/langsmith}},
  note={Accessed: 2026-02-01}
}

@misc{weave2024,
  title={Weave: Toolkit for Developing {AI}-Powered Applications},
  author={{Weights \& Biases}},
  year={2024},
  howpublished={\url{https://wandb.ai/site/weave}},
  note={Accessed: 2026-02-01}
}

@misc{braintrust2024,
  title={Braintrust: {AI} Evaluation and Observability Platform},
  author={{Braintrust Data, Inc.}},
  year={2024},
  howpublished={\url{https://www.braintrust.dev}},
  note={Accessed: 2026-02-01}
}

@misc{maxim2024,
  title={Maxim: {GenAI} Evaluation and Observability Platform},
  author={{Maxim AI}},
  year={2024},
  howpublished={\url{https://www.getmaxim.ai}},
  note={Accessed: 2026-02-01}
}

@misc{vellum2024,
  title={Vellum: {AI} Application Development Platform},
  author={{Vellum AI}},
  year={2024},
  howpublished={\url{https://www.vellum.ai}},
  note={Accessed: 2026-02-01}
}

@misc{labelstudio2024,
  title={Label Studio: Open Source Data Labeling Platform},
  author={{HumanSignal, Inc.}},
  year={2024},
  howpublished={\url{https://labelstud.io}},
  note={Accessed: 2026-01-06}
}

@misc{argilla2024,
  title={Argilla: Collaboration Tool for {AI} Engineers and Domain Experts},
  author={{Argilla, Inc.}},
  year={2024},
  howpublished={\url{https://argilla.io}},
  note={Accessed: 2026-01-06}
}

@article{chen2024critical,
  title={A Survey on Large Language Models for Critical Societal Domains: Finance, Healthcare, and Law},
  author={Chen, Zhiyu Zoey and Ma, Jing and Zhang, Xinlu and Hao, Nan and Yan, An and Nourbakhsh, Armineh and Yang, Xianjun and McAuley, Julian and Petzold, Linda and Wang, William Yang},
  journal={Transactions on Machine Learning Research},
  year={2024}
}

@article{ling2025domain,
  title={Domain Specialization as the Key to Make Large Language Models Disruptive: A Comprehensive Survey},
  author={Ling, Chen and Zhao, Xujiang and Lu, Jiaying and Deng, Chengyuan and Zheng, Can and Wang, Junxiang and Chowdhury, Tanmoy and Li, Yun and Cui, Hejie and Zhang, Xuchao and others},
  journal={ACM Computing Surveys},
  volume={58},
  number={3},
  year={2025},
  publisher={ACM},
  doi={10.1145/3764579}
}

@article{stade2024behavioral,
  title={Large Language Models Could Change the Future of Behavioral Healthcare: A Proposal for Responsible Development and Evaluation},
  author={Stade, Elizabeth C. and Wiltsey Stirman, Shannon and Ungar, Lyle H. and Boland, Cody L. and Schwartz, H. Andrew and Yaden, David B. and Sedoc, Joao and DeRubeis, Robert J. and Willer, Robb and Eichstaedt, Johannes C.},
  journal={npj Mental Health Research},
  volume={3},
  pages={12},
  year={2024},
  publisher={Nature Publishing Group},
  doi={10.1038/s44184-024-00056-z}
}

@article{katz2024bar,
  title={{GPT-4} Passes the Bar Exam},
  author={Katz, Daniel Martin and Bommarito, Michael James and Gao, Shang and Arredondo, Pablo},
  journal={Philosophical Transactions of the Royal Society A},
  volume={382},
  number={2270},
  pages={20230254},
  year={2024},
  doi={10.1098/rsta.2023.0254}
}

@article{vanveen2024clinical,
  title={Adapted Large Language Models Can Outperform Medical Experts in Clinical Text Summarization},
  author={Van Veen, Dave and Van Uden, Cara and Blankemeier, Louis and Delbrouck, Jean-Benoit and Aali, Asad and Bluethgen, Christian and Pareek, Anuj and Polacin, Malgorzata and Reis, Eduardo Pontes and Seehofnerova, Anna and others},
  journal={Nature Medicine},
  volume={30},
  pages={1134--1142},
  year={2024},
  publisher={Nature Publishing Group},
  doi={10.1038/s41591-024-02855-5}
}

@inproceedings{zamfirescu2023johnny,
  title={Why Johnny Can't Prompt: How Non-{AI} Experts Try (and Fail) to Design {LLM} Prompts},
  author={Zamfirescu-Pereira, J.D. and Wong, Richmond Y. and Hartmann, Bjoern and Yang, Qian},
  booktitle={Proceedings of the 2023 CHI Conference on Human Factors in Computing Systems},
  year={2023},
  publisher={ACM},
  doi={10.1145/3544548.3581388}
}

@article{schulhoff2024prompt,
  title={The Prompt Report: A Systematic Survey of Prompting Techniques},
  author={Schulhoff, Sander and Ilie, Michael and Balepur, Nishant and Kahadze, Konstantine and Liu, Amanda and Si, Chenglei and Li, Yinheng and Gupta, Aayush and Han, HyoJung and Schulhoff, Sevien and others},
  journal={arXiv preprint arXiv:2406.06608},
  year={2024}
}

@article{miller1956magical,
  title={The Magical Number Seven, Plus or Minus Two: Some Limits on Our Capacity for Processing Information},
  author={Miller, George A.},
  journal={Psychological Review},
  volume={63},
  number={2},
  pages={81--97},
  year={1956},
  publisher={American Psychological Association},
  doi={10.1037/h0043158}
}

@inproceedings{kiritchenko2017bws,
  title={Best-Worst Scaling More Reliable than Rating Scales: A Case Study on Sentiment Intensity Annotation},
  author={Kiritchenko, Svetlana and Mohammad, Saif},
  booktitle={Proceedings of the 55th Annual Meeting of the Association for Computational Linguistics (Volume 2: Short Papers)},
  pages={465--470},
  year={2017},
  publisher={Association for Computational Linguistics},
  doi={10.18653/v1/P17-2074}
}

\end{document}